\documentclass[letterpaper, 10 pt, journal, twoside]{IEEEtran}

\IEEEoverridecommandlockouts                              

\usepackage[top=60pt, left=48pt, right=48pt, bottom=43pt]{geometry}

\usepackage{graphicx} 
\usepackage{textcomp, amssymb}
\usepackage{amsmath}
\usepackage{gensymb}
\usepackage{makecell}
\usepackage{float}
\usepackage[font=small,labelfont=bf,tableposition=top]{caption}

\usepackage[acronym]{glossaries}
\usepackage{color}
\definecolor{dgreen}{rgb}{0,0,0}
\definecolor{dyellow}{rgb}{.7,.7,0}
\definecolor{dred}{rgb}{1,0,0}
\definecolor{dblue}{rgb}{0,0,0.7}
\definecolor{dorange}{rgb}{0.9,0.5,0.1}

\newacronym{coolname}{CALVIN}{Composing Actions from Language and Vision}
\usepackage{pifont}
\usepackage{colortbl}
\usepackage{soul}
\newcommand*\colourcheck[1]{%
  \expandafter\newcommand\csname #1check\endcsname{\textcolor{#1}{\ding{52}}}%
}
\newcommand*\colourcross[1]{%
  \expandafter\newcommand\csname #1xmark\endcsname{\textcolor{#1}{\ding{55}}}%
}
\colourcheck{green}
\colourcross{red}
\definecolor{Gray}{gray}{0.9}
\definecolor{LightCyan}{rgb}{0.88,1,1}
\makeatletter
\let\NAT@parse\undefined
\makeatother
\usepackage[hidelinks]{hyperref}
\hypersetup{
    colorlinks,
    urlcolor=magenta,
    citecolor={black},
    linkcolor={black}
    }
\usepackage{cite}
\usepackage[resetlabels]{multibib}
\newcites{New}{References}

\title{
CALVIN: A Benchmark for Language-Conditioned Policy Learning  for Long-Horizon Robot Manipulation Tasks
}

\author{Oier Mees$^{*1}$, Lukas Hermann$^{*1}$, Erick Rosete-Beas$^{1}$, Wolfram Burgard$^{2}$\\ 
 \url{http://calvin.cs.uni-freiburg.de}
 \thanks{Manuscript received: February, 23, 2022; Accepted May, 22, 2022.}
 \thanks{This paper was recommended for publication by Associate Editor S. Chernova and Editor D.
Kulic upon evaluation of the reviewers' comments.}
 \thanks{$^\ast$Equal contribution.$^{1}$University of Freiburg, Germany.$^{2}$University of Technology Nuremberg, Germany. {\tt\footnotesize meeso@informatik.uni-freiburg.de}}
\thanks{Digital Object Identifier (DOI): see top of this page.}
\thanks{© 2022 IEEE.  Personal use of this material is permitted.  Permission from IEEE must be obtained for all other uses, in any current or future media, including reprinting/republishing this material for advertising or promotional purposes, creating new collective works, for resale or redistribution to servers or lists, or reuse of any copyrighted component of this work in other works.}
}

\begin{document}

\glsunset{coolname}

\makeatletter
\let\@oldmaketitle\@maketitle
\renewcommand{\@maketitle}{\@oldmaketitle
  \includegraphics[width=\linewidth] {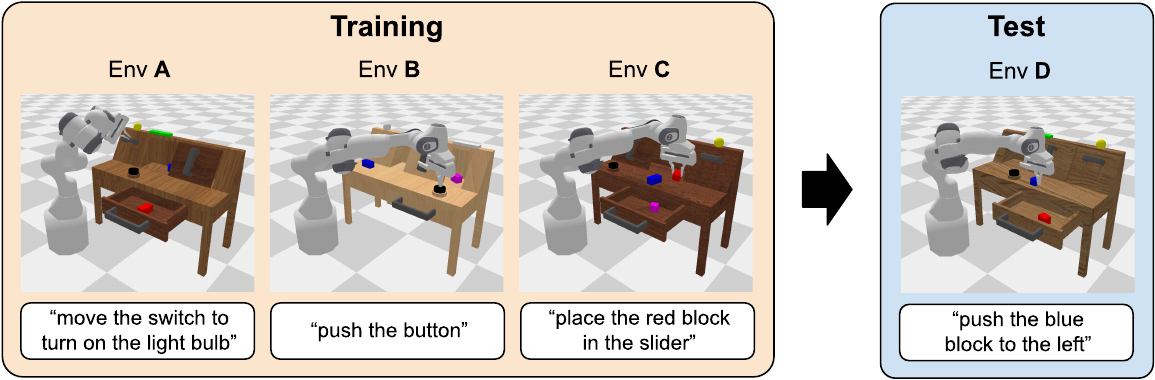} \\[0.25em]
  \refstepcounter{figure}\footnotesize{\bf{Fig. 1:} \gls{coolname} is a benchmark to learn many long-horizon language-conditioned tasks over a range of four manipulation environments, designed to be diverse yet carry shared structure, from multimodal onboard sensor observations. In the most difficult evaluation, the methods must
  generalize to unseen entities by training on a large interaction corpora covering three environments and testing on an unseen scene.}
  \label{fig:real} \medskip \vspace{-10pt}}
\makeatother

\maketitle
\markboth{IEEE Robotics and Automation Letters. Preprint Version. Accepted May, 2022}
{Mees \MakeLowercase{\textit{et al.}}: CALVIN: A Benchmark for Language-Conditioned Policy Learning for Long-Horizon Robot Manipulation Tasks} 

\renewcommand\thefigure{\arabic{figure}}
\setcounter{figure}{1}
\begin{abstract}
General-purpose robots coexisting with humans in their environment must learn to relate human language to their perceptions and actions to be useful in a range of daily tasks.
Moreover, they need to acquire a diverse repertoire of general-purpose skills that allow composing long-horizon tasks by following unconstrained language instructions.
In this paper, we present \gls{coolname} (\underline{C}omposing \underline{A}ctions from \underline{L}anguage and \underline{Vi}sio\underline{n}), an open-source simulated benchmark to learn long-horizon language-conditioned tasks. 
Our aim is to make it possible to develop agents that can solve many robotic manipulation tasks over a long horizon, from onboard sensors, and specified only via human language. \gls{coolname} tasks are more complex in terms of sequence length, action space, and language than existing vision-and-language task datasets and supports flexible specification of sensor
suites. We evaluate the agents in zero-shot to novel language instructions and to novel environments. We show that a baseline model based on multi-context imitation learning performs poorly on \gls{coolname}, suggesting that there is significant room for developing innovative agents that learn to relate human language to their world models with this benchmark.
\end{abstract}

\section{Introduction}

\IEEEPARstart{A}{long-standing} goal for robotics and embodied agents is to build systems that can perform tasks specified in natural language. Concepts expressed in natural language provide humans with an intuitive way to represent, summarize, and abstract diverse knowledge skills. By means of abstraction, concepts such as ``open the drawer and push the middle object into the drawer'' can be extended to a potentially infinite set of new and unseen entities. Additionally, humans leverage concepts to describe complex tasks as sequences of natural language instructions.
This stands in contrast to current robots, which typically lack this generalization ability and learn individual tasks one at a time.
Moreover, multi-task learning approaches traditionally assume that tasks are specified to the agent at test time via mechanisms such as goal images~\cite{lynch2020learning} and one-hot skill selectors~\cite{yu2020meta, kalashnikov2021mt} that are not practical for non-expert users to instruct robots in everyday real-world settings. 
As robots become ubiquitous across human-centered environments the need for intuitive task specification grows: how can we scale robot learning systems to autonomously acquire general-purpose knowledge that allows them to compose long-horizon tasks by following unconstrained language instructions?

To address this problem we present \gls{coolname}, a new open-source simulated benchmark that links human language to robot motor skills, behaviors, and objects in interactive visual environments. In this setting, a single agent must 
solve complex manipulation tasks by understanding a series of unconstrained language expressions in a row, e.g., ``open the drawer \ldots pick up the blue block \ldots push the block into the drawer \ldots open the sliding door''.
Furthermore, to evaluate the agents' ability for long-horizon planning, agents in this scenario are expected to be able to perform any combination of subtasks in any order. 
\gls{coolname} has been developed from the ground up to support training, prototyping, and validation of language-conditioned continuous control policies over a range of four indoor manipulation environments, visualized in Figure 1. \gls{coolname} includes ${\sim} 24$ hours teleoperated unstructured \emph{play} data together with 20K language directives. Unscripted playful interactions have the advantage of being task-agnostic, diverse, and relatively cheap to obtain~\cite{lynch2020learning, young2021playful}. The simulation platform supports a range of sensors commonly utilized for visuomotor control: RGB-D images from both a static and a gripper camera, proprioceptive information, and vision-based tactile sensing~\cite{wang2020tacto}. We believe that this flexible sensor suite will allow researchers to develop improved multimodal agents that can solve many tasks in real-world settings.
This is the first public benchmark of instruction following, to our knowledge, that combines: natural language conditioning, multimodal high-dimensional inputs, 7-DOF continuous control, and long-horizon robotic object manipulation. We provide an evaluation protocol with evaluation modes of varying difficulty by choosing different combinations of sensor suites and amounts of training environments.
This effort joins the recent efforts to standardize robotics research for better benchmarks and more reproducible results. To open the door for future development of agents that can generalize abstract concepts to unseen entities the same way humans do, we include a challenging zero-shot evaluation by training on large play corpora covering three environments and testing on an unseen scene. The language instructions used for testing are
not included in the training set and represent novel ways of describing the manipulation tasks seen during training.

To establish baseline performance levels, we evaluate the multi-context imitation learning (MCIL) approach that uses relabeled imitation learning to distill many reusable behaviors into a goal-directed policy~\cite{lynch2020language}. This model is not effective on the complex long horizon robot manipulation tasks in \gls{coolname}. While it achieves up to 53.9\% success rate in short horizon tasks, it performs poorly in the long-horizon setting.
We note that there is no constraint to use imitation learning approaches to solve \gls{coolname} tasks, as approaches that use reinforcement learning to learn language-conditioned policies can also be applied~\cite{nair2021learning}.

In summary, \gls{coolname} facilitates learning models that translate from language to sequences of motor skills in a realistic simulation environment. This benchmark captures many challenges present
in real-world settings for relating human language to robot actions and perception for accomplishing long-horizon manipulation tasks. Models that can overcome these challenges will begin to close the gap towards scalable, general-purpose, language-driven robotics.

\section{Related Work}
Natural language processing has recently received much attention in the field of robotics~\cite{tellex2020robots}, following the advances made towards learning groundings between vision and language~\cite{kazemzadeh2014referitgame, lu2019vilbert}.  Recent successes in human-robot interaction include an interactive fetching system to localize objects mentioned
in referring expressions~\cite{paul2016efficient, Shridhar-RSS-18, hatori2018interactively, nguyen2020robot, zhang2021invigorate} or grounding not only objects, but also spatial relations to follow language expressions characterizing pick-and-place commands~\cite{mees21iser, venkatesh2020spatial, liu2021structformer}. By contrast, \gls{coolname} tasks require grounding language to a wide variety of general-purpose robot skills.
Prior work on mapping language and vision to actions has been studied mostly in restricted environments~\cite{misra2017mapping, yu2018interactive} and simplified actuators with discrete motion primitives~\cite{anderson2018vision, shridhar2020alfred, shridhar2021cliport}. 
A growing body of work also looks at learning language-conditioned policies for continuous visuomotor-control in 3D environments via imitation learning~\cite{lynch2020language, stepputtis2020language, jang2021bc} or reinforcement learning~\cite{nair2021learning, shaoconcept2robot, blukis2018mapping}. These approaches typically require offline data sources of robotic interaction, such as teleoperation or autonomous exploration data, together with post-hoc crowd-sourced language labels. However, the lack of standardized benchmarks and algorithm implementations, makes it difficult to compare approaches and to facilitate future research.

The most closely related benchmark to ours is ALFRED~\cite{shridhar2020alfred}, which contains language instructions for combined navigation and manipulation tasks with seven predefined action primitives. In \gls{coolname}, rather than classifying predefined actions, the agent must learn to acquire a diverse repertoire of general-purpose skills that allows composing long-horizon tasks by following unconstrained language instructions in closed loop control.
Our tabletop environments are inspired by the one shown in Lynch \emph{et al.}~\cite{lynch2020language} in order to have a fair comparison to their MCIL approach, which we implement to establish baseline performance levels. We note that although considered a state-of-the-art approach, no public implementation of MCIL is available. In contrast to their work, \gls{coolname} contains more subtasks (34 vs 18), longer long-horizon evaluation sequences (5 vs 4), provides a range of sensors commonly utilized for visuomotor control  and allows testing zero-shot generalization by leveraging a range of four manipulation environments and unseen language instructions. Finally, CALVIN goes beyond the original MCIL setup by adding a challenging visual grounding problem, where similar language instructions for differently colored blocks are given and the agent needs to identify which block is meant.

\section{CALVIN}
The aim of the \gls{coolname} benchmark is to evaluate the learning of long-horizon language-conditioned continuous control policies. In this setting, a single agent must 
solve complex manipulation tasks by understanding a series of unconstrained language expressions in a row, e.g., ``open the drawer\ldots pick up the blue block\ldots now push the block into the drawer\ldots now open the sliding door''. 
We  note that in the benchmark we only allow feasible  sequences that can be achieved from a predefined initial environment state. The \gls{coolname} benchmark consists of three key components, which are:
\begin{enumerate}
  \item \gls{coolname} Environment
  \item \gls{coolname} Dataset
  \item \gls{coolname} Challenge
\end{enumerate}

\subsection{The CALVIN Environment}
\gls{coolname} features four different, yet structurally related environments (A, B, C, D) so that it can be used for general playing as well as evaluating specific tasks. The environments contain a 7-DOF Franka Emika Panda robot arm with a parallel gripper and a desk with a sliding door and a drawer that can be opened and closed. On the desk, there is a button that toggles a green light and a switch to control a light bulb. Besides, there are three different colored and shaped rectangular blocks. To better evaluate the generalization capabilities of the learned language groundings, all environments have different textures and all static elements such as the sliding door, the drawer, the light button, and switch are positioned differently. The position of the desk, robot, and the static camera is the same in all environments.
Due to the general difficulty of language-conditioned multi-task closed-loop control, we reduced the complexity of the objects to unicolored primitive shapes. If future advances in this field require new challenges we will reflect this by extending \gls{coolname} to environments with more realistic and diverse objects.
Physics are simulated using the PyBullet physics engine~\cite{coumans2021}, which supports fast GPU rendering for large-scale parallel data collection.
\begin{figure}[t]
    \centering
        \begin{tabular}{|c|c|}
        \hline
        \multicolumn{2}{|c|}{\textbf{Observation Space}}\\
        \hline
        RGB static camera & $200 \times 200\times 3$\\ 
        \hline
        Depth static camera & $200 \times 200$\\ 
        \hline
        RGB gripper camera & $84 \times 84\times 3$\\ 
        \hline
        Depth gripper camera & $84 \times 84$\\ 
        \hline
        Tactile image & $120 \times 160 \times 2$\\ 
        \hline
                    & EE position (3) \\
                    & EE orientation (3) \\
        Proprioceptive state & Gripper width (1) \\
                    & Joint positions (7) \\
                    & Gripper action (1) \\
        \hline
       \multicolumn{2}{|c|}{\textbf{Action Space}}\\
        \hline
        Absolute cartesian pose  & EE position (3) \\ 
        (w.r.t. world frame)         & EE orientation (3) \\
                                    & Gripper action (1) \\
        \hline
        Relative cartesian displacement  & EE position (3) \\ 
        (w.r.t. gripper frame)        & EE orientation (3) \\
                                    & Gripper action (1) \\
        \hline
        Joint action & Joint positions (7) \\
        & Gripper action (1) \\
        \hline
        \end{tabular}
    \caption{Observation and action spaces supported by \gls{coolname}.}
    \label{tab:obs_space}
\end{figure}
\subsubsection{Observation and Action Space}
\begin{figure}[t]
\centering
\setlength{\tabcolsep}{2pt}
\begin{tabular}{c c c}
\includegraphics[width=0.32\linewidth]{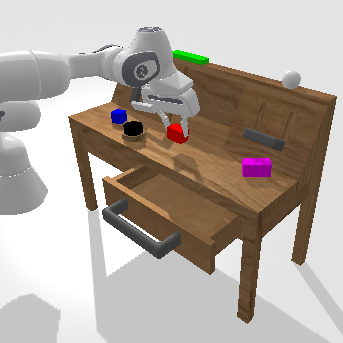}&
\includegraphics[width=0.32\linewidth]{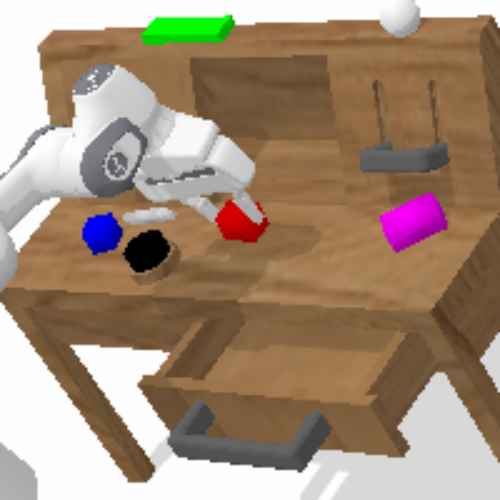}&
\includegraphics[width=0.32\linewidth]{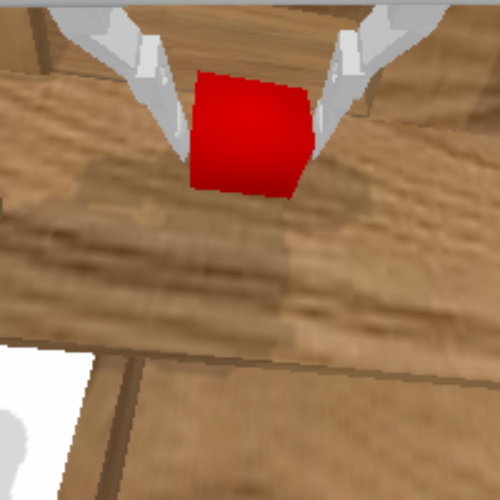}\\
\includegraphics[width=0.32\linewidth]{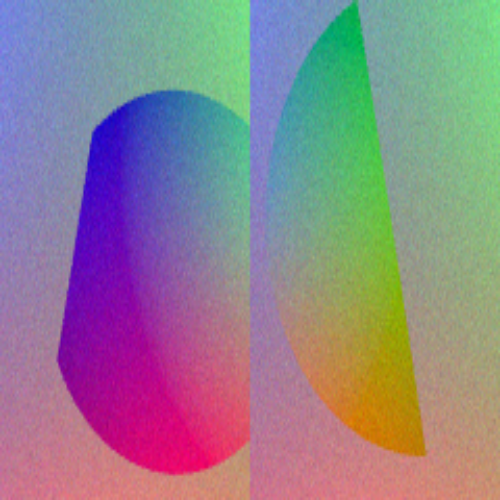}&
\includegraphics[width=0.32\linewidth]{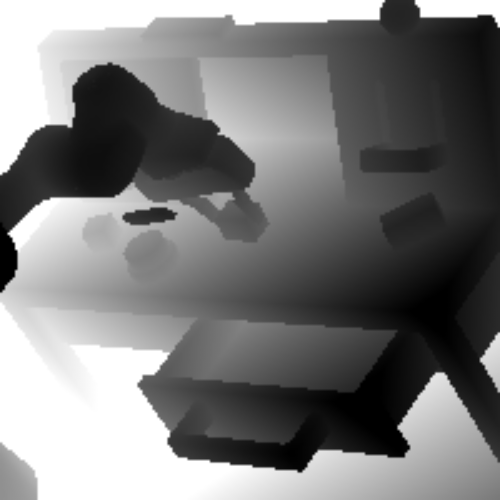}&
\includegraphics[width=0.32\linewidth]{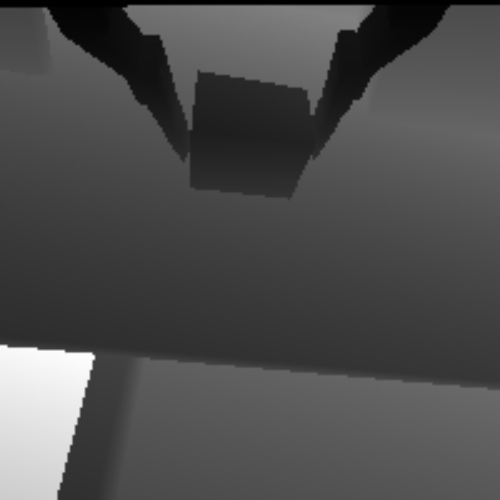}
\end{tabular}
\caption{\gls{coolname} supports a range of sensors commonly utilized for visuomotor control: RGB-D images from both a static and a gripper camera, proprioceptive information, and vision-based tactile sensing (bottom-left).}
\label{fig:sensors}
\end{figure}

Unlike prior work which relies on RGB images from an egocentric camera to perceive its surroundings~\cite{lynch2020learning, lynch2020language}, \gls{coolname} offers a range of sensors that can be used to develop and prototype agents that learn task-agnostic control in the real world. Concretely, the agent perceives its surroundings from RGB-D images from both a fixed and a gripper camera. It additionally has access to a vision-based tactile sensor~\cite{wang2020tacto} and to continuous internal proprioceptive sensors. A visualization of the supported sensor modalities is shown in Figure~\ref{fig:sensors}. The agent must perform closed-loop continuous control to follow unconstrained language instructions characterizing complex robot manipulation tasks, sending continuous actions to the robot at 30hz.
In order to give researchers and practitioners the freedom to experiment with different action spaces, \gls{coolname} supports absolute and relative cartesian actions, as well as actions in joint space.
We encourage the community to study flexible combinations of observation and action spaces since the tasks require a varying degree of precise control vs. coarse locomotion.
While the static camera and absolute cartesian actions are the natural choices for tasks that call for a complete traversal of the environment from one side to another, the gripper camera and relative actions (w.r.t to the gripper frame) allow more fine-grained control for tasks like stacking or grasping.
Tactile information can become important when the task requires the robot to maintain a stable grasp on the handle while moving the sliding door to the side. See Fig. \ref{tab:obs_space} for a description of the observation and action dimensionalities.

\subsubsection{Tasks}
We define 34 specific tasks (see Fig.~\ref{tab:tasks}) that can be achieved in each one of the environments
The environment has the functionality to automatically detect which one of the tasks has been completed in a sequence of steps, which can serve as a sparse reward for reinforcement learning agents. The criterion for task completion is defined in terms of a change in the environment state between the initial and final step of a sequence. This also enables the automatic task detection in any variable-length sequence of offline data, since the environment can be reset to the state of each one of the recorded frames.

\begin{figure}[t]
    \centering
        \begin{tabular}{|c|c|}
        \hline
        \textbf{Task} & \textbf{Natural language instructions}\\
        \hline
        rotate red block right & ``rotate the red block 90\\ 
        & degrees to the right''\\
        & ``turn the red block right''\\
        \hline 
        push blue block left & ``go slide the blue\\ 
        & block to the left''\\
        & ``push left the blue block''\\
        \hline
        move slider left & ``grasp the door handle,\\
        & then slide the door to the left''\\
        &  ``slide the door to the left''\\
        \hline
         open drawer & ``grasp the handle of the\\
         & drawer and open it''\\
         & ``go open the drawer''\\
         \hline
         lift red block & ``lift the red block\\
         & from the table''\\
         &  ``pick up the red block''\\
         \hline
         pick pink block  & ``pick up the pink\\
         from drawer & block lying in the drawer''\\
         \hline
         place in slider & ``put the grasped \\
         & object in the slider''\\
         \hline
         stack blocks & ``stack blocks on top\\
         & of each other''\\
         \hline
         unstack blocks & ``collapse the stacked blocks''\\
         & ``go to the tower of blocks \\ 
         & and take off the top one'' \\
         \hline
         turn on light bulb & ``toggle the light switch\\
         & to turn on the light bulb''\\
         \hline
         turn off green light & ``push the button to\\
         & turn off the green light''\\
         \hline
        \end{tabular}
    \caption{Example crowd-sourced natural language instructions to specify  manipulation tasks in \gls{coolname}.}
    \label{tab:tasks}
\end{figure}

\subsection{The CALVIN Dataset}
\subsubsection{Unstructured Demonstrations}
Learning generally requires exposure to diverse training data. To effectively cover state space, we collect twenty-four hours of teleoperated ``play'' data in four environments with a HTC Vive VR headset, spending an approximately equal time of six hours in each environment. This corresponds to ${\sim}$2.4M interaction steps and ${\sim}$40M short-horizon windows for relabeled goal conditioned imitation learning~\cite{kaelbling1993learning, andrychowicz2017hindsight}, each spanning 1-2 seconds. In this setting, an operator is not constrained to a set of predefined tasks, but rather engages in behavior that satisfies their own
curiosity or some other intrinsic motivation.  Unscripted playful interactions have the advantage of being task-agnostic, diverse, and relatively cheap to obtain~\cite{lynch2020learning, young2021playful}. We asked three people to collect data, and these users were untrained and given no information about the downstream tasks. The only guideline we gave data collectors was to ``explore the environment without dropping objects from the table''.
This includes picking up and placing objects, opening, and closing drawers, sliding doors, pushing buttons, operating switches and undirected actions. 
This style of data is very different from commonly used task-specific data, which only consists of expert trajectories. Playful interaction data by design is free-form, so there are no categories
associated with the data. This kind of unstructured data is useful because it contains exploratory and sub-optimal behaviors that are critical to learning generalizable and robust representations, e.g., enabling retrying behavior.
While expert demonstrations often only show one of the many possible ways to solve a task, play data is richer in the sense that it covers the multimodal space of possible solutions. 
However, as opposed to expert demonstrations, in play data some task instances naturally occur less frequently than others, especially those that have the completion of another task as a prerequisite.

\subsubsection{Language Instructions}
Approaches that learn language-conditioned continuous control policies typically require post-hoc crowd-sourced natural  language  labels aligned with its corresponding robot interaction data~\cite{lynch2020language, nair2021learning}. Instead of relying entirely on crowd-sourced annotations, we collect over 400 crowd-sourced natural language instructions corresponding to over 34 tasks and label episodes procedurally using the recorded environment state of the \gls{coolname} dataset.
We note that using this labeling scheme, only sequences that display meaningful skills are labeled with language annotations.
We visualize example language annotations in Fig.~\ref{tab:tasks}.
In order to simulate a real-world scenario where it might not be possible to pair all the collected robot experience with crowd-sourced language annotations, we annotate only 1\% of the recorded robot interaction data with language instructions. Besides language instructions, we provide precomputed language embeddings extracted from MiniLM~\cite{wang2020minilm}. MiniLM distills a large Transformer based language model and is trained on generic language corpora (e.g., Wikipedia). It has a vocabulary size of 30,522 words and maps a sentence of any length into a vector of size 384. We note that there exist many choices for encoding raw
text in a semantic pre-trained vector space and encourage the community to experiment with different choices to solve for \gls{coolname} tasks.
\begin{figure*}[t]
    \centering
    \renewcommand{\arraystretch}{1.2}
        \begin{tabular}{| c |}
        \hline
       \textbf{Long-horizon language instructions}\\
        \hline
        ``turn on the led'' $\rightarrow$ ``open drawer'' $\rightarrow$ ``push the blue blue block $\rightarrow$ ``pick up the blue block '' $\rightarrow$ ``place in slider'' \\
         \hline
        ``move slider left''  $\rightarrow$ ``lift red block from slider'' $\rightarrow$ ``stack blocks'' $\rightarrow$ ``toggle light'' $\rightarrow$ `` collapse stacked blocks'' \\
         \hline
         ``open drawer'' $\rightarrow$ ``push block in drawer'' $\rightarrow$ ``pick object from drawer'' $\rightarrow$ ``stack blocks'' $\rightarrow$ ``close drawer'' \\
         \hline

        \end{tabular}
    \caption{Example long-horizon language tasks sequences evaluated in~\gls{coolname}. We show the abbreviated subtask names instead of the full language annotations due to space constraint. }
    \label{tab:sequences}
    \vspace{-10pt}

\end{figure*}

\subsection{The CALVIN Challenge}
\gls{coolname} combines the challenging settings of open-ended robotic manipulation with open-ended human language conditioning. For example, a robot that is instructed to ``place the blue block inside the drawer'' must be able to relate language to its world model. Concretely, it needs to learn to identify how a blue block and a drawer look like in its multimodal perceptual observations\footnote{Simulator states consisting of object positions and orientations are also provided, but not used to better capture challenges of real-world settings.}, and then it needs to reason over the best sequence of actions to ``place inside the drawer''. Ideally, a general-purpose robot should be able to perform any combination of tasks instructed with natural language in any order. Thus, to accelerate progress in language-driven robotics, we present a set of evaluation protocols of varying difficulty by choosing different combinations of sensor suites and amounts of training environments.

\subsubsection{Training and Test Environments}
\gls{coolname} offers three combinations of training and test environments with varying difficulty:

\textbf{Single Environment}:
Training in a single environment and evaluating the policy in the same environment. This corresponds to the setting of Lynch \emph{et al.}~\cite{lynch2020language}.

\textbf{Multi Environment}:
Training in all four environments and evaluating the policy in one of them. This poses an additional challenge since the policy has to generalize to multiple textures and different locations of the sliding door, button, and switch. On the other hand, the agents can benefit from increased data.

\textbf{Zero-Shot Multi Environment}:
To open the door for future development of agents that can generalize abstract concepts to unseen entities the same way humans do, we include a challenging zero-shot evaluation by training in three environments and evaluating the policy in the fourth unseen one. This is the hardest combination since the policy has never seen the test environment during training. However, all elements of the scene were present in different locations in the training environments.
While highly challenging, we believe it aligns well with test-time expectations for service robots to be useful in a range of daily tasks in everyday environments. Concretely, in \gls{coolname} agents need to generalize to a room where the environment has different textures and all static elements such as the sliding door, the drawer and the light turning button and switch are positioned differently. Thus, a language-conditioned policy should ideally be able to open a sliding door even if it is differently positioned or looks visually a bit different.

\begin{figure}[h]
    \small
    \centering
    \begin{tabular}{| >{\raggedright\arraybackslash}p{0.2\linewidth} | >{\raggedright\arraybackslash}p{0.7\linewidth} |}
        \hline
        \textbf{Task} & \textbf{Condition}
        \\ \hline
        Rotate red/blue/pink block right
        & 
        The object has to be rotated clockwise more than $60^{\degree}$ around the z-axis while not being rotated more than $30^{\degree}$ around the x or y-axis.
        \\ \hline
        Rotate red/blue/pink block left 
        & 
        The object has to be rotated counterclockwise more than $60^{\degree}$ around z while not being rotated  more than $30^{\degree}$ around the x or y-axis. 
        \\\hline
        Push red/blue/pink block right
        & 
        The object has to move more than 10 cm to the right while having surface contact in both frames. 
        \\\hline
        Push red/blue/pink block left
        &
        The object has to move more than 10 cm to the left while having surface contact in both frames. 
        \\\hline
        Move slider left/right
        & 
        The sliding door has to be pushed at least 12 cm to the left/right.
        \\\hline
        Open/close drawer 
        & The drawer has to be pushed in/pulled out at least 10 cm.
        \\ \hline
        Lift red/blue/pink block table 
        & 
        The object has to be grasped from the table surface and lifted at least 5 cm high. In the first frame the gripper may not touch the object. 
        \\ \hline
        Lift red/blue/pink block slider 
        & 
        The object has to be grasped from the sliding cabinet's surface and lifted at least 3 cm. 
        In the first frame the gripper may not touch the object.
        \\ \hline
        Lift red/blue/pink block drawer
        & 
        The object has to be grasped from the drawer's surface and lifted at least 5 cm high.  
        In the first frame the gripper may not touch the object.
        \\ \hline
        Place in slider/drawer
        & 
        The object has to be placed in the sliding cabinet/drawer. It must be lifted by the gripper in the first frame.
        \\ \hline
        Push into drawer 
        & 
        The object has to be pushed into the drawer. It has to touch the table surface in the first frame.
        \\ \hline
        Stack blocks 
        & 
        A block has to be placed on top of another block. It may not be in contact with the gripper in the final frame. 
        \\ \hline
        Unstack blocks 
        &
        A block has to be removed from the top of another block. It may not be in contact with the gripper in the first frame.
        \\ \hline
        Turn on/off light bulb 
        & 
        The switch has to be pushed up/down to turn on/off the yellow light bulb.
        \\ \hline
        Turn on/off LED
        & The button has to be pressed to turn on/turn off the green LED light. 
        \\ \hline
    \end{tabular}
    \caption{List of all 34 tasks with their respective success criteria.}
    \label{tab:all_tasks}
\end{figure}

\subsubsection{Evaluation Metrics}
All three environment combinations are evaluated with the following metrics:

\textbf{Multi-Task Language Control (MTLC)}: The simplest evaluation aims to verify how well the learned multi-task language-conditioned policy generalizes to 34 manipulation tasks, which we visualize in Fig.~\ref{tab:all_tasks}. The evaluation begins by resetting the simulator to the first state of a valid unseen demonstration, to ensure that the commanded instruction is valid. For each manipulation task 10 rollouts are performed with their corresponding different starting states. The language instructions used for testing are not included in the training set and represent novel ways of describing the manipulation tasks seen during training.
\begin{figure}[h!]
    \centering
    \includegraphics[width=1.0\linewidth]{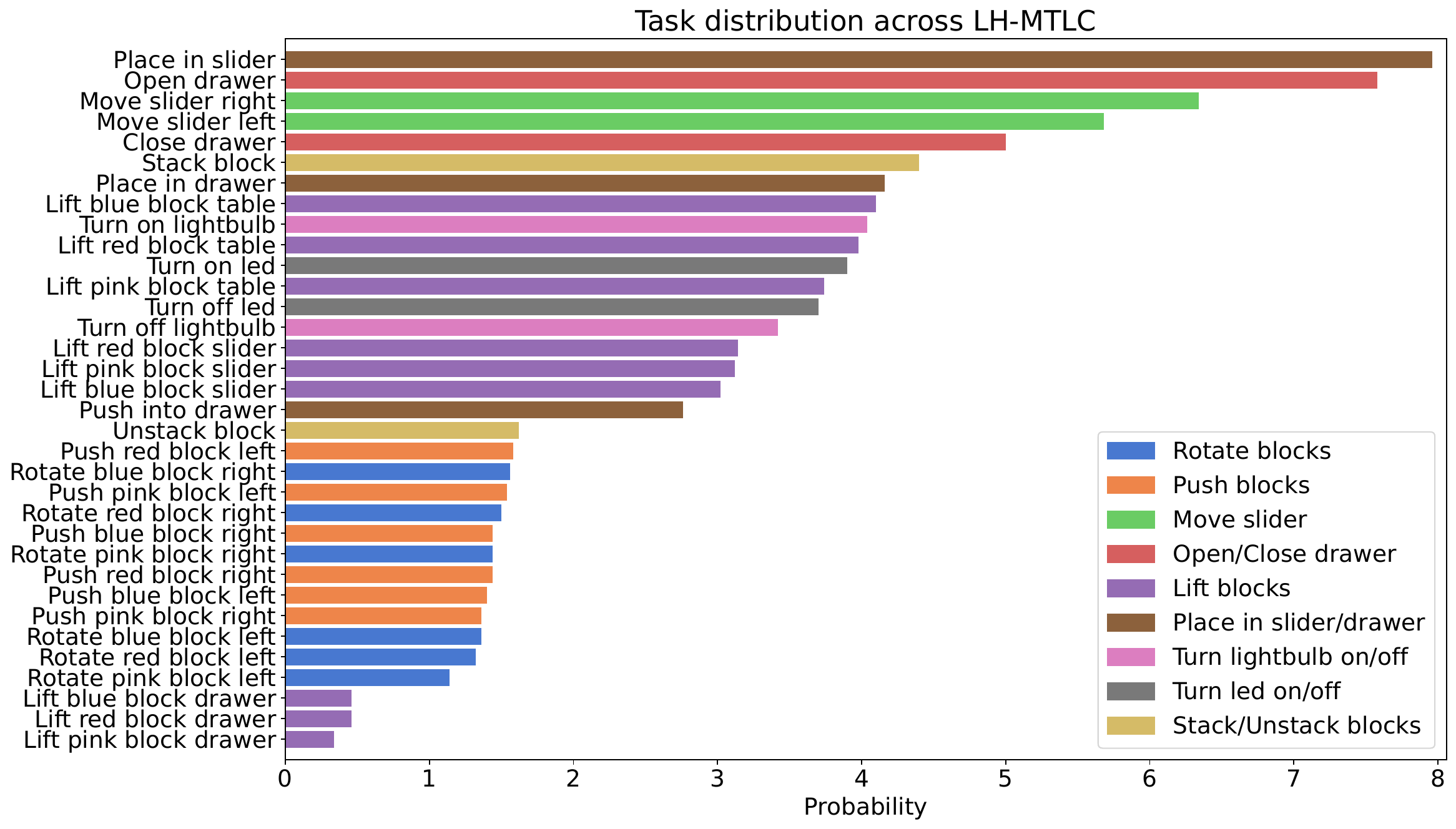}
     \caption{Visualization of the subtask distribution across the 1000 instruction chains used for the Long Horizon MTLC evaluation. We show the percentage in which each subtask appears in the distribution.}
    \label{fig:task_dist}
\end{figure}

\textbf{Long-Horizon MTLC (LH-MTLC)}: This evaluation aims to verify how well the learned  multi-task language-conditioned policy can accomplish several language instructions in a row. This setting is very challenging as it requires agents to be able to transition between different subgoals. We treat the 34 tasks of the previous evaluation as subgoals and compute valid sequences consisting of five sequential tasks. We only allow feasible sequences that can be achieved from a predefined initial environment state. We filter the evaluation sequences for cycles, redundancies and similarities to arrive at 1000 unique instruction chains. Examples for excluded sequences are ``close the drawer''\ldots ``place in drawer'' (unfeasible), ``move slider right''\ldots``move slider left''\ldots ``move slider right'' (cyclic) or ``push blue block left''\ldots``push red block left''(similar).  We reset the robot to a neutral position after every sequence to avoid biasing the policies through the robot's initial pose. We note that this neutral initialization breaks correlation between initial state and task, forcing the agent to rely entirely on language to infer and solve the task. We include different initial scene configurations in the evaluation to better evaluate generalization capabilities. We visualize the evaluated subtask distribution in Figure~\ref{fig:task_dist}. For each subtask we condition the policy on the current language instruction and transition to the next subgoal only if the agent successfully completes the current task according to the environments state indicator.
\begin{figure*}[ht]
  \centering
  \begin{tabular}{ |c| c| c| c| c| c| c| c| c| c| c| c|}
  \hline
   \multicolumn{5}{|c|}{Input} &  Train$\,\to\,$Test & MTLC & \multicolumn{5}{|c|}{LH-MTLC}\\
  \hline
        \multicolumn{2}{|c|}{Static Camera} &  \multicolumn{2}{|c|}{Gripper Camera} & Tactile & & (34 tasks) & \multicolumn{5}{|c|}{No. Instructions in a Row (1000  chains)}\\
        \cline{1-5}
        \cline{8-12}
         RGB & D & RGB & D & RGB & & & 1 & 2 & 3 & 4 & 5\\
  \hline
  
  \greencheck & \redxmark & \redxmark & \redxmark & \redxmark & D$\,\to\,$D  & 53.9\% & 48.9\%  & 12.9\% & 2.6\% & 0.5\% & 0.08\%\\
  \greencheck & \redxmark & \redxmark & \redxmark & \redxmark & A,B,C,D$\,\to\,$D  & 35.6\% & 28.2\% & 2.5\% & 0.3\% & 0\% & 0\%\\
  \greencheck & \redxmark & \redxmark & \redxmark & \redxmark & A,B,C$\,\to\,$D  & 38.6\% & 20.2\% & 0.2\% & 0\% & 0\% & 0\%\\
  
  \rowcolor{Gray}
  \greencheck & \redxmark & \greencheck & \redxmark & \redxmark & D$\,\to\,$D  & 51.8\% & 34.4\% & 5.8\% & 1.1\% & 0.2\% & 0.08\%\\
  \rowcolor{Gray}
  \greencheck & \redxmark & \greencheck & \redxmark & \redxmark & A,B,C,D$\,\to\,$D  & 49.7\% & 37.3\% & 2.7\% & 0.17\% & 0\% & 0\%\\
  \rowcolor{Gray}
  \greencheck & \redxmark & \greencheck & \redxmark & \redxmark & A,B,C$\,\to\,$D  & 38.0\% & 30.4\% & 1.3\% & 0.17\% & 0\% & 0\%\\

  \greencheck & \redxmark & \redxmark & \redxmark & \greencheck & D$\,\to\,$D  & 54.2\% & 28.5\% & 3.2\% & 0\% & 0\% & 0\%\\
  
  \greencheck & \redxmark & \redxmark & \redxmark & \greencheck & A,B,C,D$\,\to\,$D  & 47.9\% & 22.7\% & 2.3\% & 0.3\% & 0\% & 0\%\\
  
  \greencheck & \redxmark & \redxmark & \redxmark & \greencheck & A,B,C$\,\to\,$D  & 43.7\% & 17.3\% & 0.8\% & 0.08\% & 0\% & 0\%\\
  
  \rowcolor{Gray}
    \greencheck & \greencheck & \greencheck & \greencheck & \redxmark & D$\,\to\,$D  & 46.1\% & 28.2\% & 4.6\% & 0.3\% & 0.08\% & 0\%\\
  \rowcolor{Gray}
  \greencheck & \greencheck & \greencheck & \greencheck & \redxmark & A,B,C,D$\,\to\,$D  & 40.7\% & 14.4\% & 1.8\% & 0.08\% & 0.08\% & 0\%\\
  \rowcolor{Gray}
  \greencheck & \greencheck & \greencheck & \greencheck & \redxmark & A,B,C$\,\to\,$D  & 30.8\% & 21.1\% & 1.3\% & 0\% & 0\% & 0\%\\
  \hline
  \end{tabular}
  \caption{Baseline performance of  MCIL~\cite{lynch2020language} on the \gls{coolname} Challenge for different combinations of training and test environments and sensor suites. }
  \label{tab:roboexp}
\end{figure*}

\subsubsection{Sensor Combinations}
The aim of \gls{coolname} is to develop innovative agents that learn to relate human language from onboard sensors by capturing many challenges present in real-world settings.
Most autonomous robots operating in complex environments are equipped with different sensors to perceive their surroundings. To foster development and experimentation of language-conditioned policies that
perform manipulation tasks in the real-world, \gls{coolname} supports a range of sensors commonly utilized for visuomotor control: RGB-D images from both a static and a gripper camera, proprioceptive information, and vision-based tactile sensing~\cite{wang2020tacto}. We therefore evaluate baseline agents for different sensors combinations.

\section{Baseline Models}
An agent trained for \gls{coolname} needs to jointly reason over perceptual and language input and produce a sequence of low-level motor commands to interact with the environment.
\subsection{Multicontext Imitation Learning}
We model the interactive agent with a general-purpose goal-reaching policy based on multi-context imitation learning (MCIL) from play data~\cite{lynch2020language}. To learn from unstructured ``play'' we assume access to an unsegmented teleoperated play dataset $\mathcal{D}$ of semantically meaningful behaviors provided by users, without a set of predefined tasks in mind. 
To learn control, this long temporal state-action stream $\mathcal{D} = \{ (x_t, a_t )\}_{t=0}^{\infty}$ is relabeled~\cite{andrychowicz2017hindsight}, treating each visited state in the dataset as a ``reached goal state'', with the preceding states and actions treated as optimal behavior for reaching that goal. Relabeling yields a dataset of $D_{\text{play}} = \{ (\tau, x_g )_i\}_{i=0}^{D_{\text{play}}}$ where each goal state $x_g$ has a trajectory demonstration $\tau=\{ (x_0, a_0),\ldots \}$ solving for the goal. These short horizon goal image conditioned demonstrations can be fed to a simple maximum likelihood goal conditioned imitation objective:
\begin{equation}
    \mathcal{L}_{LfP} = \mathbb{E}_{(\tau, x_g) \sim D_{\text{play}}}  \left [ \sum_{t=0}^{\mid \tau \mid} \log \pi_{\theta} (a_t \mid x_t, x_g)\right ] 
\end{equation}
to learn a  goal-reaching policy $\pi_{\theta} \left(a_t \mid x_t, x_g \right)$. Multi-context imitation learning addresses the inherent multi-modality in free-form imitation datasets by auto-encoding contextual demonstrations through a latent ``plan'' space with an sequence-to-sequence conditional variational auto-encoder (seq2seq CVAE). The decoder is a policy trained to reconstruct input actions, conditioned on state $x_t$, goal $x_g$, and an inferred plan $z$ for how to get from $x_t$ to $x_g$. At test time, it takes a goal as input, and infers and follows plan $z$  in closed-loop.

However, when learning language-conditioned policies $ \pi_{\theta} \left(a_t \mid x_t, l \right)$ it is not possible to  relabel any visited state $x$ to a natural language goal as the goal space  is no longer equivalent to the observation space. Lynch \emph{et al.}~\cite{lynch2020language} showed that pairing a small number of random windows with language after-the-fact instructions enables learning a single
language-conditioned visuomotor policy that can perform a wide variety of robotic manipulation tasks. The key insight here is that solving a single imitation learning policy for either goal image or language goals, allows for learning control mostly from unlabeled play data and reduces the burden of language annotation to less than 1\% of the total data. Concretely, given multiple contextual imitation datasets $\mathcal{D} = \{D^0, D^1,\ldots ,D^K\}$, with a different way of describing
tasks,  MCIL trains a single latent goal conditioned policy $\pi_{\theta} \left(a_t \mid x_t, z \right)$ over all datasets simultaneously, as well as one parameterized encoder per dataset.

\subsection{Implementation Details}
We follow the baseline architecture implementation reported by Lynch \emph{et al.}~\cite{lynch2020language} unless stated otherwise. We train the agent with the Adam optimizer and a  learning rate of $10^{-4}$. We set the weight controlling the influence of the KL divergence to the total loss to $\beta=0.001$. During training, we randomly sample windows between length 16 and 32 and pad them until the max length of 32. As in the original implementation, no image data augmentations are applied and absolute cartesian actions w.r.t the world frame are used.
The encoder for the gripper camera takes an image of $84 \times 84$ as input and consists of 3 convolutional layers with 32, 64, and 64 channels followed by a 128 unit ReLU MLP. The encoder for the visual-tactile sensor is based on a pre-trained ResNet-18 model. The feature vectors produced by the different modality encoders are concatenated. Depth images are concatenated channel-wise with the RGB images in an early-fusion fashion.
In contrast to~\cite{lynch2020language}, the gripper fingers of the robot in the \gls{coolname} environment cannot be controlled independently, reducing the action output of the network by one dimension. We note that the same training hyperparameters are used for all splits.


\section{Experimental Results}
\label{sec:result}
The results comparing language-conditioned policies based on multicontext imitation learning for the different evaluation modes in \gls{coolname} are shown in Figure~\ref{tab:roboexp}. We note that there is no constraint to use imitation learning approaches to solve \gls{coolname} tasks, as approaches that use reinforcement learning to learn language-conditioned policies can also be applied~\cite{nair2021learning}. We observe that the baseline with images of the static camera  achieves a success rate of  53.9\% for the MTLC evaluation setting, when training and testing the 34 manipulation tasks on the same environment. The success rate stays comparable when including a gripper camera,  depth channels or tactile sensing.
We hypothesize that the reason for not seeing larger improvements when adding the gripper camera is that the policy might benefit from using relative actions instead of global actions. A qualitative analysis indicates that the performance depends significantly on the initial position of the robot, suggesting the agent relies on context rather than learning to disentangle initial states and tasks. It is possible this is due to causal confusion between the proprioceptive information and the target actions~\cite{de2019causal}. Besides, we did not use image data augmentations in the baselines to stay close to the original implementation, but we hypothesize this might be beneficial. 
Additionally, more elaborate sensor fusion approaches such as mixture of experts~\cite{mees2016choosing, lee2019making} or view-invariant contrastive learning~\cite{tian2020contrastive, mees20icra_asn}  might be necessary to learn better multimodal state representations. 

For the Long-Horizon MTLC evaluation we observe that the agents perform poorly on \gls{coolname}'s long-horizon tasks with high-dimensional state spaces. The best MCIL model achieves a success rate of 0.08\% when following chains of five language instructions in a row when training and testing on the same environment. Additionally, it solves the first subtask of the chain, starting from a neutral position, in 48.9\% of the cases. We observe that the policy sometimes correctly executes block manipulation tasks, but confuses the red and blue block colors in the instruction. As the language models embed sentences containing the words red and blue similarly, backpropagating through the entire language model and leveraging auxiliary losses that try to align visual and language representations~\cite{radford2021learning} might be beneficial to tackle the complicated perceptual grounding problem. 

Finally, the general performance drops significantly when evaluating on the multi environment and zero-shot multi environment settings, which do not follow the standard assumption of imitation learning that training and test tasks are drawn independently from the same distribution. In order to achieve better zero-shot generalization capabilities, additional techniques from the domain adaptation literature~\cite{mees20icra_asn}, better data augmentation and a stronger focus on depth inputs, since they are invariant to texture changes, might be helpful. As MCIL is an offline learning method, we hypothesize that na\"{i}ve data sharing between multiple domains can be brittle because it can exacerbate the distribution shift between the policy represented in the data and the policy being learned~\cite{yu2021conservative}. 
This motivates further research into agents that can perform the complex long-horizon language-conditioned manipulation tasks introduced by \gls{coolname}.


\section{Conclusion}
\label{sec:conclusion}

In this paper, we presented \gls{coolname}, the first public benchmark of instruction following that combines natural language conditioning, multimodal high-dimensional inputs, 7-DOF continuous control, and long-horizon robotic object manipulation in both seen and unseen environments. As the field of language-driven robotics evolves, a need arises to standardize research for better benchmarks and more reproducible results. \gls{coolname} has the goal of providing researchers with a modular framework that has been developed from the ground up to support training, prototyping, and validation of language-conditioned continuous control policies. Further to that, we hope, along with the help of the community, to continuously expand the tasks available for both training and evaluation. 

We use \gls{coolname} to evaluate a conditional sequence-to-sequence variational autoencoder, shown to be effective in other long horizon language-conditioned manipulation tasks~\cite{lynch2020language}.
While this model is relatively competent at accomplishing some subgoals, the overall success rates are poor. The long horizon of \gls{coolname} tasks poses a significant challenge with sub-problems including the acquisition of a diverse repertoire  of general-purpose skills, object detection, referring expression and action grounding, and task-agnostic continuous control. 
We hope \gls{coolname} will open the door for the future development of agents that can relate human language to their perception and actions and generalize abstract concepts to unseen entities in the same way humans do.







\section*{Acknowledgement}
This work was supported by the German Federal Ministry of Education and Research under contract 01IS18040B-OML. We thank Corey Lynch and Pierre Sermanet for help with the MCIL baseline.

\bibliographystyle{IEEEtran}
\bibliography{root}

\clearpage
\markboth{}{}

\onecolumn
\appendix
\subsection{Tasks}
All tasks are defined in terms of change in the environment state between the first and the final frame of a sequence.
In order to see if a task was solved in an arbitrary sequence of frames of the CALVIN dataset,
the environment is reset to the state of the first and the last frame of that sequence.
The tasks detector compares the two simulator states and checks which task conditions are fulfilled. A key advantage of this strategy is that it enables efficient evaluation of sequences for task completion independent of their length.
Figure \ref{tab:all_tasks2} shows a list of all task definitions.

\begin{figure}[h]
    \centering
    \def\arraystretch{1.2}
        \begin{tabular}{| l | l | } 
        \hline
        \textbf{Task} & \textbf{Condition}\\
        
        \hline
        \makecell[l]{Rotate red block right \\ Rotate blue block right \\ Rotate pink block right} 
        & \makecell[l]{The object has to be rotated clockwise more than $60^{\degree}$ around the z-axis \\while not being rotated for more than $30^{\degree}$ around the x or y-axis.} \\
        
        \hline
        \makecell[l]{Rotate red block left \\ Rotate blue block left \\ Rotate pink block left} 
        & \makecell[l]{The object has to be rotated counterclockwise more than $60^{\degree}$ around the z-axis \\while not being rotated for more than $30^{\degree}$ around the x or y-axis.} \\
        
        \hline
        \makecell[l]{Push red block right \\ Push blue block right \\ Push pink block right} 
        & \makecell[l]{The object has to move more than 10 cm to the right while having surface  \\ 
        contact in both frames} \\
        
        \hline
        \makecell[l]{Push red block left \\ Push blue block left \\ Push pink block left} 
        & \makecell[l]{The object has to move more than 10 cm to the left while having surface  \\ 
        contact in both frames} \\
        
        \hline
        \makecell[l]{Move slider left} 
        & \makecell[l]{The sliding door has to be pushed at least 12 cm to the left.} \\
        
        \hline
        \makecell[l]{Move slider right} 
        & \makecell[l]{The sliding door has to be pushed at least 12 cm to the right.} \\
        
        \hline
        \makecell[l]{Open drawer} 
        & \makecell[l]{The drawer has to pulled out at least 10 cm.} \\
        
        \hline
        \makecell[l]{Close drawer} 
        & \makecell[l]{The drawer has to be pushed in at least 10 cm.} \\
        
        \hline
        \makecell[l]{Lift red block table \\ Lift blue block table \\ Lift pink block table} 
        & \makecell[l]{The object has to be grasped from the table surface and lifted at least 5 cm high. \\ 
        In the first frame the gripper may not touch the object.} \\
        
        \hline
        \makecell[l]{Lift red block slider \\ Lift blue block slider \\ Lift pink block slider} 
        & \makecell[l]{The object has to be grasped from the surface of the sliding cabinet and lifted  \\ at least 3 cm high. 
        In the first frame the gripper may not touch the object.} \\
        
        \hline
        \makecell[l]{Lift red block drawer \\ Lift blue block drawer \\ Lift pink block drawer} 
        & \makecell[l]{The object has to be grasped from the surface of the drawer and lifted at least 5 cm high. \\ 
        In the first frame the gripper may not touch the object.} \\
        
        \hline
        \makecell[l]{Place in slider} 
        & \makecell[l]{The object has to be placed in the sliding cabinet.  \\ It must be lifted by the gripper in the first frame.} \\
        
        \hline
        \makecell[l]{Place in drawer} 
        & \makecell[l]{The object has to be placed in the drawer. \\ It must be lifted by the gripper in the first frame.} \\
        
        \hline
        \makecell[l]{Push into drawer} 
        & \makecell[l]{The object has to be pushed into the drawer. \\ It has to touch the table surface in the first frame.} \\

        \hline
        \makecell[l]{Stack blocks} 
        & \makecell[l]{A block has to be placed on top of another block.  \\It may not be in contact with the gripper in the final frame. } \\
        
        \hline
        \makecell[l]{Unstack blocks} 
        & \makecell[l]{A block that is stacked on another block has to be removed from the top, \\ either by grasping it or by pushing it down. It may not be in contact \\ with the gripper in the first frame. } \\
        
        \hline
        \makecell[l]{Turn on light bulb} 
        & \makecell[l]{The switch has to be pushed down to turn on the yellow light bulb. } \\
        
        \hline
        \makecell[l]{Turn off light bulb} 
        & \makecell[l]{The switch has to be pushed up to turn off the yellow light bulb. } \\
        
        \hline
        \makecell[l]{Turn on LED} 
        & \makecell[l]{The button has to be pressed to turn on the green LED light.} \\
        
        \hline
        \makecell[l]{Turn off LED} 
        & \makecell[l]{The button has to be pressed to turn off the green LED light.} \\
        
        \hline
        
        \end{tabular}
    \caption{List of all 34 tasks with their respective success criteria.}
    \label{tab:all_tasks2}
\end{figure}

\subsection{Language Annotation Generation}
The language annotations are extracted automatically from the recorded data with the following procedure: we randomly sample sequences with a window size of 64 frames. 
For each sequence the task detector checks if a task has been solved between the first and the last frame. 
Additionally, we check that neither that task nor any other task is solved in the first half of the sequence.
The intuition behind this is that we want to include the locomotion behavior prior to the actual task. For example, before opening the drawer, the arm must navigate in the direction of the handle. This is important for learning to solve tasks with language goals from arbitrary starting positions. 
If a sequence qualifies for labeling, we sample a natural language instruction from a set of predefined sentences with approximately 11 synonymous instructions per task. In total, this gives 389 unique language instructions for 34 tasks.
The sequence in which the task ``stack blocks'' is solved could for example get instructions such as ``place the grasped block on top of another block'' or ``stack blocks on top of each other''.
In  order  to  simulate  a  real-world  scenario  where  it  might not  be  possible  to  pair  all  the  collected  robot  experience with crowd-sourced language annotations, we annotate only 1\%  of  the  recorded  robot  interaction  data  with  language instructions. 
The CALVIN dataset conveniently includes precomputed MiniLM language embeddings for all instructions, but researchers are free to use their own language model of choice on the raw input data.

\end{document}